\documentclass[11pt]{article}

\usepackage[final]{acl}

\usepackage{times}
\usepackage{latexsym}

\usepackage[T1]{fontenc}

\usepackage[utf8]{inputenc}

\usepackage{microtype}

\usepackage{inconsolata}

\usepackage{graphicx}

\usepackage{amsmath}
\usepackage{amssymb}
\usepackage{multirow}
\usepackage{booktabs}
\usepackage{amsmath}
\usepackage{tikz}
\usepackage[none]{hyphenat}
\usepackage{placeins}
\usepackage{caption}
\usepackage{algorithm}
\usepackage{algorithmic}

\title{Training LLMs Beyond Next Token Prediction - Filling the \\ Mutual Information Gap}

\author{
 \textbf{Chun-Hao Yang\textsuperscript{1}\textsuperscript{*}},
 \textbf{Bo-Han Feng\textsuperscript{1}\textsuperscript{*}},
 \textbf{Tzu-Yuan Lai\textsuperscript{1}},
 \\
 \textbf{Yan-Yu Chen\textsuperscript{1}},
 \textbf{Yin-Kai Dean Huang\textsuperscript{1}},
 \textbf{Shou-De Lin\textsuperscript{1,2}}
\\
 National Taiwan University\textsuperscript{1} \\
 National Taiwan University AI Center of Research Excellence\textsuperscript{2}
\\
 \small
 \texttt{b10902139@csie.ntu.edu.tw, b10902031@csie.ntu.edu.tw, frogcute122609@gmail.com} \\
 \small
 \texttt{b10902137@csie.ntu.edu.tw, dean.huang.tw@gmail.com, sdlin@csie.ntu.edu.tw}
}

\begin{document}
\maketitle
\def\thefootnote{*}\footnotetext{Equal contribution.}\def\thefootnote{\arabic{footnote}}
\begin{abstract}
Autoregressive language models are usually trained with next-token prediction under a fixed left-to-right target order. This paper studies target-token organization as a training-time design choice across arithmetic reasoning, multi-label text classification, and text generation. Controlled order studies show that the default NTP order can be far from the best-performing tested organization, while fixed hand-crafted orders such as reverse order do not generalize across tasks. To address this, \textbf{Max(MI($S;t$))} is introduced as a deterministic mutual-information-based heuristic that prioritizes information-rich target tokens. For fixed-length outputs, it constructs task-specific prediction orders; for open-vocabulary text generation, the same principle is applied through lightweight prefix augmentation. Experiments with GPT-style models and pretrained Llama/Qwen variants show that Max(MI($S;t$)) often improves over standard NTP and simple baselines such as reverse order, TF-IDF, and conditional-probability selection, without changing model architecture or autoregressive decoding.
\end{abstract}

\section{Introduction}

\begin{figure*}[t]
\centering
\includegraphics[width=0.98\linewidth]{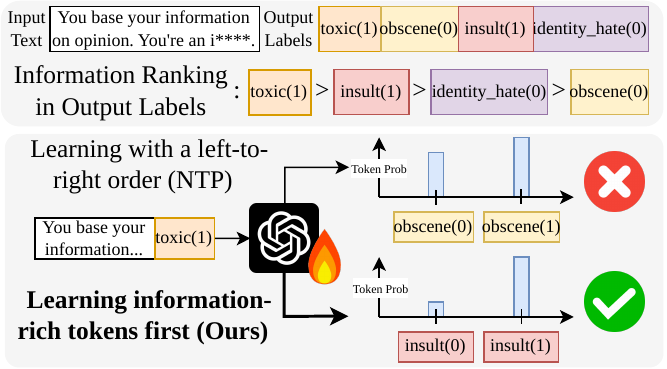}
\caption{An example of learning information-rich tokens first in MLC task. \textbf{Prioritizing information-rich tokens} (\textit{insult} label) during training reduces prediction uncertainty and improves subsequent label prediction, whereas \textbf{left-to-right NTP} can be harder to learn due to a lack of emphasis on informative tokens and cumulative uncertainty.}
\label{fig:dataformat}
\end{figure*}

Large language models (LLMs) can be improved at both test time and training time. Prior work has shown that LLMs are sensitive to the ordering of information presented at test time. Such order effects have been observed across few-shot learning \citep{kumar2021reorderingexampleshelpsprimingbased, lu2022fantasticallyorderedpromptsthem}, reasoning premises \citep{chen2024premiseordermattersreasoning}, answer options in multiple-choice evaluation \citep{pezeshkpour2023largelanguagemodelssensitivity, gupta2024changinganswerorderdecrease}, and retrieved or long-context evidence in RAG-style settings \citep{liu-etal-2024-lost, cao2025evaluatingretrievalrobustnesslarge}. These findings suggest that sequence organization is not merely a formatting detail, but a factor that shapes how autoregressive models use information. However, existing studies primarily manipulate the input context at test time. The analogous question at training time remains underexplored: under a fixed model architecture and standard autoregressive decoding, can learning be improved by organizing the prediction target more effectively before training begins?

Next-token prediction (NTP) is the dominant training objective for LLMs, but it also fixes target prediction to a left-to-right order. This default order may be suboptimal for tasks with latent structure or strong inter-token dependencies.

In this paper, we consider three types of tasks: multi-step arithmetic computations \citep{shen2023positionaldescriptionmatterstransformers}, multi-label classification (MLC) of text \citep{zhang2021enhancinglabelcorrelationfeedback, jaggi2024accuratedataefficienttoxicityprediction, zhang-etal-2024-message}, and long-form text generation (TG) \citep{schmidt2019generalizationgenerationcloserlook,he2021exposurebiasversusselfrecovery}. In these settings, errors on earlier target tokens may propagate and accumulate. In arithmetic, self-attention must implicitly learn operations such as carry propagation without explicit structural constraints. In MLC, early incorrect labels can bias later predictions toward frequent co-occurrences rather than true label relationships. In TG, early token choices can degrade subsequent coherence, and exposure bias may further amplify this effect. These observations suggest that the default left-to-right order is not always the best target organization for learning.

Take the MLC task in Figure \ref{fig:dataformat} as an example. If an LLM is trained to output information-rich labels first (\textit{insult} label before \textit{obscene} label), it tends to have lower uncertainty when predicting those tokens. Correctly predicting such high-information tokens can, in turn, provide stronger guidance for subsequent predictions. In contrast, NTP, a strict left-to-right generation objective, is more difficult for LLMs to learn, as it does not explicitly prioritize informative tokens and may propagate uncertainty step by step.

\begin{figure*}[t]
\includegraphics[width=0.98\linewidth]{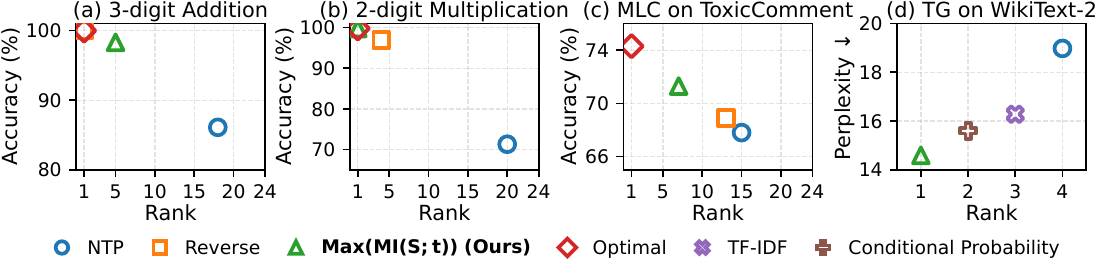}
  \caption{Rank-based comparison of target-token organization strategies. For (a)--(c), we exhaustively evaluate all $4! = 24$ orders for four output digits or labels; NTP ranks poorly among these orders, while Max(MI($S;t$)) consistently approaches or matches the optimum, and reverse order only works well for addition. For (d), where reverse ordering does not apply to natural language generation, we instead expose one selected word earlier by prefix duplication; Max(MI($S;t$)) achieves the lowest perplexity, outperforming NTP, TF-IDF, and conditional-probability baselines. For experimental setup, see Appendix B. For separate illustrative training-curve experiments showing performance discrepancy across training orders, see Figure \ref{fig:curves}.}
  \label{fig:curves_summarized}
\end{figure*}

The work most closely related to ours addresses this limitation of NTP in arithmetic and shows that reversing the output order can help small transformers on addition tasks \citep{lee2023teachingarithmeticsmalltransformers, nogueira2021investigating}. While insightful, reverse order is still a hand-crafted, task-specific strategy motivated by the standard algorithm of addition, and it does not provide a unified principle for choosing prediction orders across different arithmetic tasks or more general NLP tasks.

To address the suboptimality of NTP and the limited generality of fixed reverse ordering, we propose \textbf{Max(MI($S;t$))}, a deterministic strategy prioritizing information-rich target tokens according to their Mutual Information (MI) with the source and previously selected targets. For arithmetic and MLC, where the number of output tokens is defined in advance, \textbf{Max(MI($S;t$))} iteratively selects the most informative remaining target token to construct a task-specific prediction order. For TG, it selects one information-rich word for each sentence and duplicates it as a prefix through preprocessing-time augmentation. Both implementations follow the same principle: expose information-rich target content earlier during training. Importantly, our approach modifies only the training target organization with model architecture and standard autoregressive inference unchanged.

Figure \ref{fig:curves_summarized} shows that no single fixed order generalizes across four representative settings. For arithmetic and MLC in Figure \ref{fig:curves_summarized} (a)--(c), where each output contains four target digits or labels, we exhaustively evaluate all $4! = 24$ prediction orders. The default NTP order is often far from optimal, ranking 18th, 20th, and 15th in 3-digit addition, 2-digit multiplication, and MLC, respectively. Reverse order matches the best order for addition, but its advantage quickly diminishes in multiplication and MLC, where it ranks 4th and 13th. These results suggest that the reverse order is task-dependent, not a general solution. In contrast, proposed information-based ordering \textbf{Max(MI($S;t$))} remains consistently competitive with 5th, 1st, and 7th ranks in the same three settings, achieving accuracies close to the corresponding optima. 

For TG in Figure \ref{fig:curves_summarized} (d), where exhaustive target permutation and reverse ordering are not applicable, we apply \textbf{Max(MI($S;t$))} by a lightweight augmentation that duplicates one selected information-rich word earlier in each sentence. This strategy achieves the lowest perplexity among all baselines, reducing perplexity from 18.98 under standard NTP to 14.59, and outperforming TF-IDF and conditional-probability-based baselines with the same word selection and prefix duplication procedure. Together, these results indicate that target organization is a substantial training-time factor, and that prioritizing information-rich content offers a more adaptive alternative to fixed left-to-right or hand-crafted reverse ordering.

\paragraph{Contributions.} \textbf{1. Problem formulation.} We frame target-token organization as a training-time design choice for autoregressive language modeling, showing that left-to-right NTP can be far from the best tested order and that fixed reverse order does not reliably generalize beyond addition. \textbf{2. Method.} We propose \textbf{Max(MI($S;t$))}, a deterministic MI-based heuristic that prioritizes target tokens most informative with respect to the source and previously selected targets, yielding task-specific prediction orders without changing the model architecture, training objective, or decoding procedure. \textbf{3. Open-vocabulary extension.} We extend the same principle to text generation and summarization via a lightweight preprocessing augmentation that exposes selected information-rich words earlier in each sentence. \textbf{4. Evaluation and practicality.} Across arithmetic, multi-label classification, WikiText-2 text generation, and XSUM summarization with GPT-style, Llama, and Qwen models, \textbf{Max(MI($S;t$))} achieves stronger aggregate performance than NTP, reverse order, TF-IDF, and conditional-probability baselines, with only preprocessing-time changes.

\section{Related Work}

Prior work has shown that LLMs are sensitive to the ordering of information at inference time. In few-shot learning, changing the order of demonstrations can lead to large performance differences \citep{kumar2021reorderingexampleshelpsprimingbased, lu2022fantasticallyorderedpromptsthem}. Similar effects have been observed for answer options in multiple-choice question answering \citep{pezeshkpour2023largelanguagemodelssensitivity, gupta2024changinganswerorderdecrease}, premises in reasoning tasks \citep{chen2024premiseordermattersreasoning}, and digit order in arithmetic prompting \citep{singh2024tokenizationcountsimpacttokenization}. Order and position effects also appear in long-context and RAG settings, where models may fail to use evidence depending on where it appears or how retrieved documents are ordered \citep{liu-etal-2024-lost, cao2025evaluatingretrievalrobustnesslarge}. These studies show that sequence organization affects how autoregressive models use information, but they mainly manipulate the input context at test time.

The work most closely related to our studies targets reordering during arithmetic training \citep{lee2023teachingarithmeticsmalltransformers, nogueira2021investigating}. It shows that reversing the output order can improve small transformers on addition, suggesting that the standard left-to-right target order imposed by next-token prediction may be suboptimal. However, reverse order is a hand-crafted strategy motivated by the standard addition algorithm, and its generality beyond addition remains unclear. In contrast, our work studies target organization as a broader training-time design choice across arithmetic, MLC, and TG.

\section{Information-Rich Token Prediction}

Not all target tokens are equally informative during training. Some tokens are more strongly determined by the source, and once predicted, can provide useful context for remaining targets. Instead of always following a fixed left-to-right order, our goal is to predict target tokens sharing the most information with the current source earlier. We formalize this idea with MI.

\begin{figure}[t]
\includegraphics[width=0.98\linewidth]{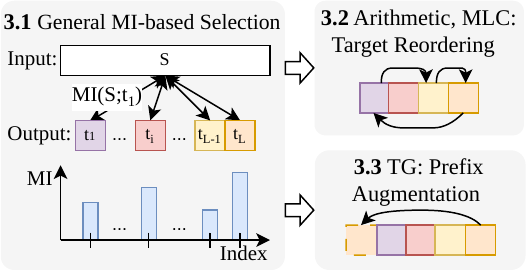}
  \caption{Overview of \textbf{Max(MI($S;t$))}. We first compute MI scores between the current source and candidate targets. Fixed-length tasks use the selected targets to form a reordered output sequence, while open-vocabulary TG uses the selected word for per-sentence prefix augmentation.}
  \label{fig:method}
\end{figure}

Figure \ref{fig:method} summarizes the method flow. We first score candidate targets by their MI with the current source. For fixed-length outputs such as arithmetic digits and MLC labels, the highest-MI target is appended to the source and the process iterates to produce a reordered target sequence. For open-vocabulary TG, the same criterion selects one information-rich word per sentence for prefix duplication. The following subsections define the general MI-based selection rule and its two instantiations.

\subsection{General MI-based Selection}

We introduce \textbf{Max(MI($S;t$))}, a deterministic strategy that constructs a prediction order by iteratively selecting the remaining target position with the highest MI with the current source. Let
$S^{(0)} = [s_1, \cdots, s_N] \in \mathbb{R}^{L_1 \times N}$ denote the initial source sequences, where $N$ is the number of training samples, $L_1$ is the source length, and each column $s_i$ is the source sequence of the $i^{\text{th}}$ sample. Let
$T = [t_1, \cdots, t_{L_2}] \in \mathbb{R}^{N \times L_2}$ denote the target tokens, where $L_2$ is the target length, $t_j$ collects the $j^{\text{th}}$ target token across all samples, and $t_{ji}$ is the $j^{\text{th}}$ target token of sample $i$.

At selection step $k$, let $S^{(k)}$ be the current source after appending the $k$ previously selected target positions to the original source, and let $s_i^{(k)}$ be its value for sample $i$. For each remaining target position $j$, let $V_j$ be the set of possible values of $t_{ji}$. We compute:
\begin{equation}
\begin{aligned}
    &\text{MI}(S^{(k)};t_j) = \\&
    \sum_{s_i^{(k)} \in S^{(k)}} \sum_{t_{ji} \in V_j}
    P(s_i^{(k)}, t_{ji})
    \log \frac{P(s_i^{(k)}, t_{ji})}{P(s_i^{(k)})P(t_{ji})}.
    \label{eq:mi_fixed}
\end{aligned}
\end{equation}

The next target position is selected as:
\begin{equation}
    j^* = \arg\max_{j \in R^{(k)}} \text{MI}(S^{(k)};t_j),
    \label{eq:mi_select}
\end{equation}
where $R^{(k)}$ is the set of remaining target positions. The selected target $t_{j^*}$ is appended to the source to form $S^{(k+1)}$ and removed from $R^{(k)}$. This process repeats until all target positions are ordered. Further theoretical justification is provided in Appendix \ref{sec:A.1}, and a step-by-step calculation example is given in Appendix \ref{sec:A.2}.

\subsection{Fixed-Length Target Reordering}

Arithmetic and MLC have fixed-length targets, making explicit target reordering feasible. Arithmetic tasks predict output digits, while MLC tasks predict a sequence of label decisions. For a 2-digit multiplication task
($A_1A_2 \times B_1B_2 = C_1C_2C_3C_4$), the source is $A_1A_2B_1B_2$ and the candidate target positions are $C_1, C_2, C_3, C_4$. If
\begin{equation}
    \arg\max_j \text{MI}(\textit{source}; C_j) = 3,
    \label{eq:arith_select}
\end{equation}
then $C_3$ is predicted first and added to the source for the next selection step. The same procedure applies to MLC by replacing digit positions with label positions: the label with the highest MI with the input text is predicted first, then used as source for selecting subsequent labels. Algorithm \ref{algo:info_rich} in Appendix \ref{sec:algo} summarizes this procedure formally.

For arithmetic and MLC, MI is computed from empirical counts over a small discrete target space. Thus, preprocessing only requires an offline counting-and-sorting pass on a CPU and does not require training an auxiliary model.

\subsection{Open-Vocabulary Prefix Augmentation} \label{sec:tg_method}

For text generation, target sentences have variable length and open-vocabulary tokens, making exhaustive target reordering impractical. We therefore apply the same information-rich-token principle through preprocessing-time augmentation: for each target sentence, we select one information-rich word and duplicate it at the beginning of the sentence.

Let $S=(s_1,s_2,\cdots,s_{N_s})$ denote the source context before the target sentence, where $N_s$ is the source-context length, and let $T=(t_1,t_2,\cdots,t_{M})$ denote the words in the target sentence, where $M$ is the target-sentence length. Unlike arithmetic and MLC, where $P(s_i,t_j)$ can be estimated directly from finite target spaces, TG involves open-vocabulary text. To approximate MI, we assume a first-order Markov structure where the probability of a target word depends primarily on the previous context token. We train a two-layer logistic regression classifier over token bigrams to estimate the joint probability of observing a target token $t_j$ after a context token $s_i$.

Let $\mathbf{e}_{s_i} \in \mathbb{R}^d$ and $\mathbf{e}_{t_j} \in \mathbb{R}^d$ be the embedding vectors for tokens $s_i$ and $t_j$, respectively. We parameterize the estimated joint probability as:
\begin{equation}
    P(s_i, t_j) =
    \sigma\left(
    \mathbf{w}_2
    \left(
    \mathbf{w}_1[\mathbf{e}_{s_i};\mathbf{e}_{t_j}] + b_1
    \right) + b_2
    \right),
    \label{eq:lr1}
\end{equation}
where $\mathbf{w}_1 \in \mathbb{R}^{h \times 2d}$, $\mathbf{w}_2 \in \mathbb{R}^{1 \times h}$, $b_1 \in \mathbb{R}^h$, and $b_2 \in \mathbb{R}$ are trainable parameters with hidden size $h$, and $\sigma$ is the sigmoid function. The model is trained using adjacent token pairs from the training corpus.

For each candidate word $t_j$ in the target sentence $T$, we define:
\begin{equation}
    \text{MI}(S;t_j)
    =
    P(S,t_j)
    \log
    \frac{P(S,t_j)}
    {P(S)P(t_j)}.
    \label{eq:5}
\end{equation}

Under the Markov approximation, $P(S,t_j)$ is estimated using the final source-context token $s_{N_s}$, yielding the ranking score:
\begin{equation}
    \text{MI-Score}(t_j)
    =
    P(s_{N_s},t_j)
    \log
    \frac{P(s_{N_s},t_j)}
    {P(t_j)} ,
    \label{eq:6}
\end{equation}
where $P(t_j)$ is the marginal probability of token $t_j$, obtained by aggregating its joint probabilities with possible context tokens in the corpus. We then select:
\begin{equation}
    t^* = \arg\max_{t_j \in T} \text{MI-Score}(t_j).
    \label{eq:lr3}
\end{equation}

For practical preprocessing cost in TG, MI scoring requires a lightweight auxiliary model because the target space is open-vocabulary. This makes preprocessing more expensive than in arithmetic and MLC, but it remains a one-time offline step; see Appendix~\ref{sec:B.3} for complexity analysis and empirical runtime.

\begin{table*}[t]
\centering
\resizebox{\linewidth}{!}{
\begin{tabular}{llcccc}
    \toprule
    \textbf{Task} &
    \textbf{Best Order} &
    \textbf{Best Acc.} &
    \textbf{NTP} &
    \textbf{Reverse} &
    \textbf{Max(MI($S$;$t$))}
    \\\toprule
    3-digit Addition                & $C_4C_3C_2C_1$ & 99.99\% & 85.09\% & \textbf{99.99\%} & 97.46\% \\\midrule
    2-digit Multiplication          & $C_3C_2C_4C_1$ & 98.22\% & 70.15\% & 95.35\% & \textbf{97.38\%} \\\midrule
    4-digit Logarithm               & $C_1C_4C_3C_2$ & 98.21\% & 95.99\% & \textbf{96.63\%}$^\dagger$ & \textbf{96.63\%}$^\dagger$ \\\midrule
    3-digit Greatest Common Divisor & $C_1C_3C_2$ & 95.80\% & 91.45\% & \textbf{92.59\%}$^\dagger$ & \textbf{92.59\%}$^\dagger$ \\\midrule
    3-digit Multiplication          & $C_5C_4C_3C_6C_2C_1$ & 99.66\% & 71.43\% & 88.15\% & \textbf{96.23\%} \\\midrule
    2-digit Chicken-Rabbit Problem  & $C_4C_2C_3C_1$ & 92.12\% & 34.92\% & 45.04\% & \textbf{89.48\%} \\\midrule
    \textbf{Average}                &   & 97.33\% & 74.84\% & 86.29\% & \textbf{94.96\%} \\\bottomrule
\end{tabular}}
\caption{
GPT-based model performance across different arithmetic tasks. Token prediction strategies: (1) \textbf{NTP}: standard left-to-right target order. (2) \textbf{Reverse}: reversed NTP. (3) \textbf{Max(MI($S$;$t$))}: prioritizes output digits with the highest mutual information with the source. $^\dagger$: Identical predicted token sequences for both strategies.
}
\label{tab:arithmetic_res_gpt}
\end{table*}

\section{Experiments for Arithmetic Tasks}

Arithmetic tasks provide controlled settings for studying target-token organization: output digits have fixed positions, all target-order permutations are enumerable for small tasks, and exact correctness can be measured. We evaluate \textbf{Max(MI($S$;$t$))} against \textbf{NTP}, the standard left-to-right target order, and \textbf{Reverse}, the reversed NTP order. For GPT-based models, we exhaustively test all target-token permutations to identify the best tested order. We further evaluate the same strategies by fine-tuning a billion-parameter Qwen-Math model in both multi-task and long-sequence multiplication settings. Experimental details are provided in Appendix \ref{sec:B.1}.

\begin{figure}[H]
\centering
\includegraphics[width=0.42\textwidth]{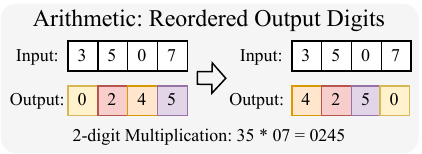}
\caption{Arithmetic data format with reordered digits. The source expression is fixed, while output digits are rearranged according to the selected prediction order.}
\label{fig:arithmetic_dataformat}
\end{figure}

\subsection{Datasets \& Evaluation}
We consider addition, multiplication, logarithm, greatest common divisor (GCD), and the chicken-rabbit (CR) problem. As shown in Figure~\ref{fig:arithmetic_dataformat}, each example consists of a fixed source expression and reordered output digits. For GPT-based experiments, each task is programmatically generated and trained separately; the number of target-order permutations ranges from $3!$ for 3-digit outputs to $6!$ for 6-digit outputs.

For example, $\log_{10}(1234)=3.091$ is represented as $[1,2,3,4,3,0,9,1]$, with target digits $C_1C_2C_3C_4=[3,0,9,1]$. For Qwen-Math fine-tuning, we mix all arithmetic tasks used in the GPT-based experiments except 3-digit multiplication, and keep mathematical symbols to match instruction-style arithmetic data, e.g., $379+841=1270$ is represented as $[3,7,9,+,8,4,1,=,1,2,7,0]$.

We also evaluate long-sequence multiplication with Qwen-Math by multiplying two $d$-digit numbers for $d=3,\ldots,8$, where the output length increases up to 16 digits. Due to the rapidly growing data space, we sample 100,000 examples for each long-sequence setting.

Predicted target orders are converted back to numerical answers. We report fixed-iteration accuracy, where the iteration count is determined by the early plateau of the best-performing order.

\subsection{Results}
Table \ref{tab:arithmetic_res_gpt} shows that target-token order substantially affects arithmetic learning. \textbf{Max(MI($S$;$t$))} achieves the highest average accuracy among the three practical strategies, reaching 94.96\% compared with 74.84\% for NTP and 86.29\% for Reverse. It also approaches the exhaustive-search upper bound of 97.33\%, indicating that neither left-to-right nor reverse order is universally reliable across arithmetic tasks.

\begin{table}[H]
\centering
    \resizebox{\linewidth}{!}{
    \begin{tabular}{lccc}\toprule
        \textbf{Task} & \textbf{NTP} & \textbf{Reverse} & \textbf{Max(MI($S$;$t$))}
        \\\midrule
        3-digit Addition                      & 99.3\% & 99.8\% & \textbf{100.0\%} \\\midrule
        2-digit Multiplication                & 94.1\% & 95.4\% & \textbf{96.7\%} \\\midrule
        4-digit Logarithm                     & 89.1\% & 89.0\%$^\dagger$ & \textbf{89.6\%}$^\dagger$ \\\midrule
        3-digit Greatest Common Divisor       & 94.0\% & 93.7\%$^\dagger$ & \textbf{94.3\%}$^\dagger$ \\\midrule
        2-digit Chicken-Rabbit Problem        & 84.7\% & 96.9\% & \textbf{99.4\%} \\\midrule
        Average                               & 92.2\% & 95.0\% & \textbf{96.0\%} \\
        \bottomrule
    \end{tabular}
    }
    \caption{
       Qwen-Math performance across arithmetic tasks. \textbf{NTP}: standard left-to-right order; \textbf{Reverse}: reversed NTP order; \textbf{Max(MI($S$;$t$))}: prioritizes information-rich output digits. $^\dagger$: Identical predicted token sequences for Reverse and Max(MI($S$;$t$)).
    }
    \label{tab:arithmetic_res_qwen}
\end{table}

Table \ref{tab:arithmetic_res_qwen} shows the same trend for multi-task Qwen-Math fine-tuning. Max(MI($S$;$t$)) achieves the best average accuracy, exceeding NTP by 3.8\% and Reverse by 1.0\%. Along with the GPT-based exhaustive studies, this suggests Max(MI($S$;$t$)) as a task-adaptive alternative to fixed hand-crafted orders.

\begin{figure}[H]
\centering
\includegraphics[width=0.48\textwidth]{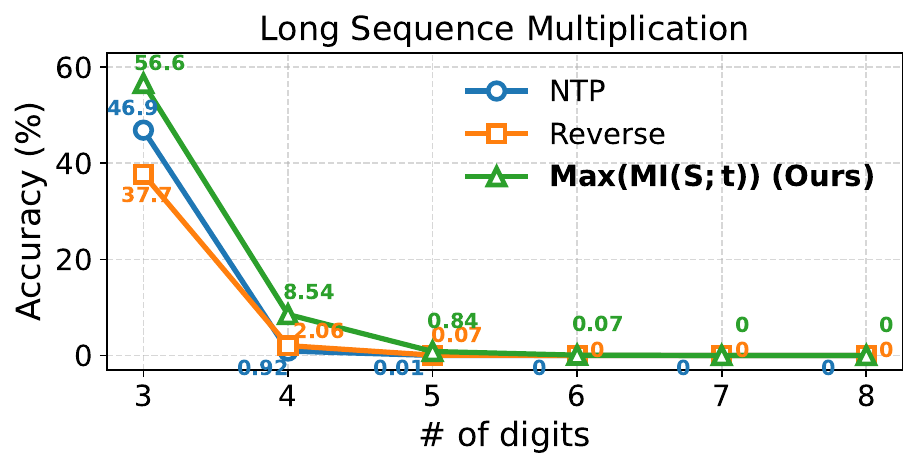}
\caption{
Long-sequence multiplication with Qwen-Math. Each task multiplies two $d$-digit numbers. All strategies degrade as output length grows, but Max(MI($S$;$t$)) shows better robustness before the longest settings collapse.
}
\label{fig:mul_long_seq}
\end{figure}

Figure \ref{fig:mul_long_seq} further evaluates longer multiplication outputs. All strategies degrade sharply as digit length increases and fail in the longest settings, but Max(MI($S$;$t$)) maintains higher accuracy than NTP and Reverse before this collapse, indicating better robustness under increasing output length.

\section{Experiments for Multi-label Classification}

\begin{table*}[t]
\centering
    \resizebox{\linewidth}{!}{
    \begin{tabular}{lllcccc}
        \toprule
        \textbf{Model} &
        \textbf{Dataset} &
        \textbf{Best Order} &
        \textbf{Best Acc.} &
        \textbf{NTP} &
        \textbf{Reverse} &
        \textbf{Max(MI($S$;$t$))}
        \\\toprule
        Llama-3.1-8B-Instruct  & ToxicComment        & $C_3C_1C_2C_4$ & 85.81\% & 84.22\% & 82.59\% & \textbf{85.14\%} \\\midrule
        Llama-3.2-1B-Instruct  & ToxicComment        & $C_1C_4C_3C_2$ & 77.20\% & 74.65\% & 74.48\% & \textbf{75.04\%} \\\midrule
        Qwen2.5-3B-Instruct    & ToxicComment        & $C_1C_3C_4C_2$ & 78.85\% & 76.24\% & 70.98\% & \textbf{78.76\%} \\\midrule
        Qwen2.5-1.5B-Instruct  & ToxicComment        & $C_2C_1C_3C_4$ & 74.26\% & 67.87\% & 68.90\% & \textbf{71.96\%} \\\midrule
        GPT-2                  & ToxicComment        & $C_3C_1C_2C_4$ & 88.63\% & 88.02\% & \textbf{88.17\%} & 88.15\% \\\midrule
        GPT-2                  & PaperAbstract       & $C_4C_1C_2C_3$ & 88.73\% & \textbf{88.52\%}$^\dagger$ & 88.23\% & \textbf{88.52\%}$^\dagger$ \\\midrule
        GPT-2                  & GoEmotions          & $C_2C_3C_4C_1$ & 96.40\% & \textbf{96.24\%}$^\dagger$ & 96.18\% & \textbf{96.24\%}$^\dagger$ \\\midrule
        Average                &                     &                & 84.27\% & 82.25\% & 81.36\% & \textbf{83.40\%}  \\\bottomrule
    \end{tabular}
    }
    \caption{
        Maximal accuracy in MLC across models and English datasets. \textbf{NTP}: standard left-to-right label order; \textbf{Reverse}: reversed NTP order; \textbf{Max(MI($S$;$t$))}: prioritizes information-rich label tokens. $^\dagger$ indicates identical predicted label sequences.
    }
\label{tab:classification_res}
\end{table*}

MLC provides a structured NLP setting for studying target-token organization: each input text is mapped to a fixed set of label decisions, and label tokens carry semantic information. Unlike multiple-choice question-answering options such as A/B/C/D, MLC labels can express both text-label relevance and dependencies among labels. We evaluate \textbf{Max(MI($S$;$t$))} against \textbf{NTP} and \textbf{Reverse} across multiple datasets, languages, and model families. Experimental details are provided in Appendix \ref{sec:B.2}.

\subsection{Datasets \& Evaluation}
\begin{figure}[t]
\centering
\includegraphics[width=0.42\textwidth]{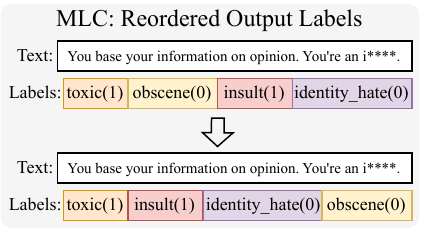}
\caption{MLC data format with reordered labels. The input text is fixed, while output label decisions are rearranged according to the selected prediction order.}
\label{fig:mlc_dataformat}
\end{figure}

We evaluate on three MLC datasets: \emph{ToxicComment}, \emph{PaperAbstract}, and \emph{GoEmotions}. To control target length, we retain the four most frequent labels in each dataset and denote them as $C_1,C_2,C_3,C_4$. As shown in Figure \ref{fig:mlc_dataformat}, each example keeps the input text fixed and reorders only the output label decisions.

For multilingual evaluation, we additionally translate \emph{ToxicComment} into Chinese using Google Translate. This setting tests whether target-order effects persist when the model may have different pretraining exposure to the input language. We report label-level accuracy.

\subsection{Results}
Table \ref{tab:classification_res} shows that target-label order affects MLC performance across models and English datasets. On average, \textbf{Max(MI($S$;$t$))} achieves 83.40\% accuracy, outperforming NTP by 1.15\% and Reverse by 2.04\%. Reverse order is weaker than NTP on average, contrasting with its strong performance in arithmetic and suggesting that fixed hand-crafted orders do not transfer reliably across task types.

\begin{table}[H]
\centering
    \resizebox{\linewidth}{!}{
    \begin{tabular}{lccc}
        \toprule
        \textbf{Model} & \textbf{NTP} & \textbf{Reverse} & \textbf{Max(MI($S$;$t$))}
        \\\midrule
        Llama-3.1-8B$^\dagger$        & 81.52\% & 80.16\% & \textbf{82.82\%}  \\\midrule
        Llama-3.2-1B$^\dagger$        & 72.84\% & 72.39\% & \textbf{74.61\%}  \\\midrule
        Qwen2.5-3B$^\dagger$          & 72.73\% & 71.92\% & \textbf{74.39\%}  \\\midrule
        Qwen2.5-1.5B$^\dagger$        & 68.08\% & \textbf{70.39\%} & 70.18\%  \\\midrule
        GPT-2               & 75.65\% & 69.42\% & \textbf{79.70\%} \\\midrule
        Average             & 74.16\% & 72.86\% & \textbf{76.34\%}   \\
        \bottomrule
    \end{tabular}
    }
    \caption{
        Maximal accuracy in MLC on the Chinese \emph{ToxicComment} dataset. \textbf{NTP} denotes the standard left-to-right label order; \textbf{Reverse} denotes the reversed NTP order; \textbf{Max(MI($S$;$t$))} prioritizes information-rich label tokens. $^\dagger$ indicates instruction-tuned models.
    }
\label{tab:classification_res_zh}
\end{table}

Table \ref{tab:classification_res_zh} shows the same comparison on Chinese \emph{ToxicComment}. Max(MI($S$;$t$)) again obtains the best average accuracy, improving over NTP by 2.18\% and over Reverse by 3.48\%. The gains are larger than in the English setting, suggesting that target-token organization may interact with language and pretraining exposure. Overall, MLC results support the need for task-adaptive label ordering rather than a universal fixed order.

\section{Experiments for Text Generation}

For open-vocabulary text generation, exhaustive target reordering is impractical. We therefore evaluate the prefix-augmentation strategy introduced in Section \ref{sec:tg_method}: for each sentence, one selected word is duplicated at the beginning of the sentence. We compare four strategies: \textbf{NTP}, the standard unaugmented baseline; \textbf{TF-IDF}, which selects the word with the highest TF-IDF score; \textbf{P($t|S$)}, which selects the word with the highest conditional probability given the source; and \textbf{Max(MI($S$;$t$))}, which selects the most information-rich word. We evaluate these strategies on WikiText-2 continued pretraining using perplexity and on XSUM summarization using ROUGE. Experimental details are provided in Appendix \ref{sec:B.3}, and additional NLU results are reported in Appendix \ref{sec:C}.



\begin{figure}[H]
\centering
\includegraphics[width=0.42\textwidth]{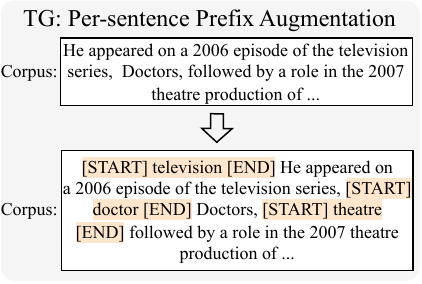}
\caption{TG data format with per-sentence prefix augmentation. The selected word is inserted before each sentence between \texttt{[START]} and \texttt{[END]}.}
\label{fig:tg_dataformat}
\end{figure}

\subsection{Datasets \& Evaluation}

For text generation, GPT-2 and Llama-3.2-1B are continually pretrained on \emph{WikiText-2}. We tokenize and lemmatize the corpus, segment documents into sentences using punctuation, select one word per sentence according to each strategy, map the selected word back to its surface form, and duplicate it at the sentence beginning with \texttt{[START]} and \texttt{[END]} delimiters, as shown in Figure \ref{fig:tg_dataformat}. We evaluate using perplexity. To compare original and augmented data on the same evaluable tokens, we mask the inserted special tokens, selected words, and the immediately following token during perplexity computation.

For summarization, Llama-3.2-1B-Instruct is fine-tuned on \emph{XSUM}. For all augmented strategies, we concatenate document and summary, apply the same per-sentence word selection and prefix augmentation as WikiText-2, and then split the augmented text back into document and summary fields for fine-tuning. We evaluate generated summaries using ROUGE-1, ROUGE-2, and ROUGE-L, with \texttt{[START]}, \texttt{[END]}, and inserted words removed for fair comparison.

\begin{table}[t]
\centering
    \resizebox{\linewidth}{!}{
    \begin{tabular}{lcccc}
        \toprule
        \textbf{Model} & \textbf{NTP} & \textbf{TF-IDF} & \textbf{P($t|S$)} & \textbf{Max(MI($S$;$t$))}
        \\\midrule
        GPT-2            & 18.98 & 16.28 & 15.60 & \textbf{14.59} \\\midrule
        Llama-3.2-1B     & 8.19  & 7.51  & 7.27  & \textbf{6.95}  \\\midrule
        Average          & 13.58 & 11.90 & 11.44 & \textbf{10.77} \\
        \bottomrule
    \end{tabular}
    }
    \caption{
       Perplexity on WikiText-2. Lower is better. NTP denotes the standard unaugmented baseline; TF-IDF, P($t|S$), and Max(MI($S$;$t$)) use the same prefix-augmentation format with different word-selection criteria.
    }
\label{tab:generation_res}
\end{table}

\begin{table}[t]
\centering
    \resizebox{\linewidth}{!}{
    \begin{tabular}{lcccc}
        \toprule
        \textbf{Metric} & \textbf{NTP} & \textbf{TF-IDF} & \textbf{P($t|S$)} & \textbf{Max(MI($S$;$t$))}
        \\\midrule
         ROUGE-1 & 0.2838 & 0.3298 & 0.3301 & \textbf{0.3313} \\\midrule
         ROUGE-2 & 0.0862 & 0.1093 & 0.1091 & \textbf{0.1102} \\\midrule
         ROUGE-L & 0.2127 & 0.2547 & 0.2534 & \textbf{0.2554} \\\midrule
         Average & 0.1942 & 0.2313 & 0.2309 & \textbf{0.2323} \\
        \bottomrule
    \end{tabular}
    }
    \caption{
       ROUGE scores on XSUM using Llama-3.2-1B-Instruct. Higher is better.
    }
    \label{tab:summarization_res}
\end{table}

\subsection{Results}

Table \ref{tab:generation_res} shows that prefix augmentation reduces WikiText-2 perplexity for both GPT-2 and Llama-3.2-1B. Max(MI($S$;$t$)) performs best, lowering average perplexity from 13.58 under NTP to 10.77, while TF-IDF and P($t|S$) also outperform NTP. This suggests that earlier exposure to selected words benefits continued pretraining, with the MI-based criterion being the strongest strategy.

Table \ref{tab:summarization_res} shows a similar trend on XSUM. Max(MI($S$;$t$)) achieves the best ROUGE-1/2/L and average ROUGE scores, though only modestly ahead of TF-IDF and P($t|S$). Overall, information-rich prefix augmentation improves open-vocabulary generation without architectural or decoding changes.

\section{Conclusion}
We studied target-token organization as a training-time design choice for autoregressive language modeling. Across arithmetic, MLC, and TG, left-to-right NTP can be suboptimal, while fixed hand-crafted orders, such as reverse order, do not consistently generalize. Max(MI($S$;$t$)) prioritizes information-rich target content via fixed-length reordering or open-vocabulary prefix augmentation. Experiments with GPT-style models and pretrained Llama/Qwen variants show that it often improves over NTP and simple baselines, highlighting target organization as a practical direction for LLM training.

\section*{Limitations}

Our study has several limitations. First, Max(MI($S$;$t$)) is a heuristic for target organization and does not guarantee the optimal prediction order. In fixed-length tasks, the best order can still depend on task-specific structure, such as positional dependencies in arithmetic. The long-sequence multiplication results further show that target organization alone does not solve long-output arithmetic scaling when sequences become very long. In open-vocabulary TG, our method uses prefix augmentation rather than exhaustive reordering, so it only approximates the broader space of possible target organizations.

Second, our experiments focus on decoder-only models from the GPT, Llama, and Qwen families. Although the results cover both models trained from scratch and pretrained variants, the effect of target organization may vary with model scale, architecture, pretraining data, and tokenizer design. Further validation is needed for larger models and other architectures.

Third, the multilingual MLC results suggest that language and pretraining exposure may affect the benefit of target organization. However, our Chinese setting is based on translated ToxicComment data, so it does not fully isolate linguistic effects from translation artifacts or domain differences. Broader multilingual evaluation is needed to better understand these factors.

Finally, Max(MI($S$;$t$)) requires estimating MI between source and target tokens. For arithmetic and MLC, this is inexpensive because the target space is small and discrete; for open-vocabulary TG, it requires an auxiliary offline scoring model. Although the preprocessing cost is a one-time step, more scalable MI approximations would be useful for larger corpora and longer contexts.

\section*{Acknowledgments}

This material is based upon work supported by National Science and Technology Council, ROC under grant number 114-2221-E-002-134-MY3, NTU AI Center of Research Excellence within Taiwan Centers of Excellence in Artificial Intelligence, and by National Taiwan University and Academia Sinica Innovative Joint Program, under grant AS-NTU-114-06. We acknowledge the use of ChatGPT for assisting in the grammatical review and refinement of this document.


\bibliography{custom}

\appendix

\section{Details of Information-Rich Token Prediction Method}

\subsection{Theoretical Justification}\label{sec:A.1}

\paragraph{Assumption.} The LLM predictions can be modeled as a Markov chain $I \rightarrow \tilde{T} \rightarrow T$, where $I$ is the input token, $\tilde{T}$ is the model output token, and $T$ is the ground-truth token. This reflects the autoregressive nature of LLMs, where the next token is sampled conditionally.

\paragraph{Connection to MI.} Minimizing mean-square error (MSE) between model predictions $\tilde{T}$ and ground-truth tokens $T$ is equivalent to maximizing the $\text{MI}(T; \tilde{T})$:
\begin{equation}
\min_{T} \mathbb{E}[||T - \tilde{T}||^2] \iff \max \text{MI}(T; \tilde{T})
\label{eq:mse_mi}
\end{equation}
\textbf{Proof Sketch.} This equivalence was demonstrated in \citet{jing2022retrievalbasedtimeseries}, where the prediction objective under a Gaussian assumption leads to maximal MI between the predicted and true values. Intuitively, higher MI reduces uncertainty and improves reconstruction accuracy.

\paragraph{Motivation via DPI.} Given the Markov chain $I \rightarrow \tilde{T} \rightarrow T$, and using the Data Processing Inequality (DPI), we obtain:
\begin{equation}
\text{MI}(I; T) \leq \text{MI}(\tilde{T}; T)
\label{eq:dpi}
\end{equation}
Hence, MI($I$;$T$) can be viewed as a proxy for how much source information is available for predicting the target. This motivates prioritizing target tokens with higher MI with the source.

\paragraph{Implication.} During training, prioritizing the target token $t$ with maximum MI with the source input $I$, i.e., $\arg\max_t \mathrm{MI}(I;t)$, encourages the model to predict more source-informative targets earlier.

\subsection{Calculation Example of MI($S$;$t_j$)} \label{sec:A.2}

We provide a step-by-step demonstration of MI($S$;$t_{j}$) calculation to clarify our method. Assume there are three 2-digit addition data samples:
\begin{itemize}
    \item $12 + 34 = 046$
    \item $55 + 66 = 121$
    \item $07 + 08 = 015$
\end{itemize}

$S$ would be:

\begin{equation}
\begin{bmatrix} s_1, s_2, s_3 \end{bmatrix} =
\begin{bmatrix}
\begin{pmatrix} 1 \\ 2 \\ 3 \\ 4\end{pmatrix}
\begin{pmatrix} 5 \\ 5 \\ 6 \\ 6\end{pmatrix}
\begin{pmatrix} 0 \\ 7 \\ 0 \\ 8\end{pmatrix}
\end{bmatrix}
\label{eq:mi_cal_s}
\end{equation}

In equation \ref{eq:mi_cal_s}, $s_i$ is the $i^{th}$ data sample with $L_1 = 4$ and $N = 3$.

$T$ would be:

\begin{equation}
\begin{aligned}
\begin{bmatrix} t_1, t_2, t_3 \end{bmatrix} &=
\begin{bmatrix}
\begin{pmatrix} t_{11} \\ t_{12} \\ t_{13} \end{pmatrix}
\begin{pmatrix} t_{21} \\ t_{22} \\ t_{23} \end{pmatrix}
\begin{pmatrix} t_{31} \\ t_{32} \\ t_{33} \end{pmatrix}
\end{bmatrix} \\&=
\begin{bmatrix}
\begin{pmatrix} 0 \\ 1 \\ 0 \end{pmatrix}
\begin{pmatrix} 4 \\ 2 \\ 1 \end{pmatrix}
\begin{pmatrix} 6 \\ 1 \\ 5 \end{pmatrix}
\end{bmatrix}
\end{aligned}
\label{eq:mi_cal_t}
\end{equation}

In equation \ref{eq:mi_cal_t}, $t_{ji}$ is $j^{th}$ target token of $i^{th}$ data sample with $N = 3$ and $L_2 = 3$. $V_j$ for $t_j$ is $\{t_{j1},\ t_{j2},\ t_{j3}\}$.

Therefore, to calculate MI($S$; $t_1$), which is MI($source$, $C_1$) in Algorithm \ref{algo:info_rich}, we can obtain the value of $\text{MI}(S;t_1) = \sum_{s_i \in S} \sum_{t_{1i} \in V_1} P(s_i, t_{1i}) \log \frac{P(s_i, t_{1i})}{P(s_i) P(t_{1i})}$ from the summation of the following terms:

\begin{itemize}
    \item $P(s_i = \begin{pmatrix} 1 \\ 2 \\ 3 \\ 4 \end{pmatrix}, t_{1i} = 0) \log(...)$, \\
    \item $P(s_i = \begin{pmatrix} 1 \\ 2 \\ 3 \\ 4 \end{pmatrix}, t_{1i} = 1) \log(...)$, \\
    \item $P(s_i = \begin{pmatrix} 5 \\ 5 \\ 6 \\ 6 \end{pmatrix}, t_{1i} = 0) \log(...)$, \\
    \item $P(s_i = \begin{pmatrix} 5 \\ 5 \\ 6 \\ 6 \end{pmatrix}, t_{1i} = 1) \log(...)$, \\
    \item $P(s_i = \begin{pmatrix} 0 \\ 7 \\ 0 \\ 8 \end{pmatrix}, t_{1i} = 0) \log(...)$, \\
    \item $P(s_i = \begin{pmatrix} 0 \\ 7 \\ 0 \\ 8 \end{pmatrix}, t_{1i} = 1) \log(...)$.
\end{itemize}

\subsection{Pseudocode for Information-Rich Token Selection} \label{sec:algo}

\begin{algorithm}[H]
\caption{Information-Rich Token Selection}\label{algo:info_rich}
\begin{flushleft}
\textbf{Input}: source tokens $[A_1, A_2, B_1, B_2]$, target tokens $[C_1, C_2, C_3, C_4]$\\
\textbf{Output}: reordered token sequence
\end{flushleft}
\begin{algorithmic}[1]
\STATE $source \gets [A_1, A_2, B_1, B_2]$
\STATE $target \gets [C_1, C_2, C_3, C_4]$
\STATE $final \gets source$
\WHILE{$target \neq \emptyset$}
    \STATE $x \gets \arg\max_{C \in target} \text{MI}(source; C)$
    \STATE Append $x$ to $source$ and $final$
    \STATE Remove $x$ from $target$
\ENDWHILE
\STATE \textbf{return} $final$
\end{algorithmic}
\end{algorithm}

\section{Experimental Setup}\label{sec:B}

\subsection{Arithmetic Tasks}\label{sec:B.1}

For the main arithmetic experiments reported in Section 4, we use a minGPT\footnote{\url{https://github.com/karpathy/mingpt}} model trained from scratch. The model is configured with a context size of $32$, using $6$ attention heads, $6$ transformer blocks, and $192$ embedding dimension. Training is performed with the AdamW optimizer, a learning rate of $5 \times 10^{-5}$, batch size $64$, and weight decay $0.1$. All runs are trained for up to $5 \times 10^{6}$, iterations with early stopping based on validation accuracy. Experiments are run on five NVIDIA RTX 3090 GPUs. The environment includes PyTorch 2.7.1, Hugging Face Transformers 4.5, CUDA 12.9, and Python 3.10.18 on Ubuntu 20.04.6 LTS.

\begin{table}[h]
\centering
\resizebox{\linewidth}{!}{
\begin{tabular}{ll}
    \toprule
    \textbf{Task Name} & \textbf{Data Format}
    \\\toprule
    \multirow{2}{*}{3-digit Addition}                & $A_1A_2A_3 + B_1B_2B_3$ \\
                                                     & $ = C_1C_2C_3C_4$
                                                     \\\midrule
    \multirow{2}{*}{2-digit Multiplication}          & $A_1A_2 \times B_1B_2$  \\
                                                     & $ = C_1C_2C_3C_4$
                                                     \\\midrule
    \multirow{2}{*}{4-digit Logarithm}               & $\log_{10} {(A_1A_2A_3A_4)}$ \\  
                                                     & $ = C_1.C_2C_3C_4$
                                                     \\\midrule
    \multirow{2}{*}{3-digit GCD}                     & $gcd(A_1A_2A_3,B_1B_2B_3)$ \\
                                                     & $ = C_1C_2C_3$
                                                     \\\midrule
    \multirow{2}{*}{3-digit Multiplication}          & $A_1A_2A_3 \times B_1B_2B_3$ \\
                                                     & $ = C_1C_2C_3C_4C_5C_6$
                                                     \\\midrule
    \multirow{2}{*}{2-digit CR}                      & $ChickenRabbit(A_1A_2, B_1B_2)$ \\
                                                     & $ = C_1C_2C_3C_4$
                                                     \\\midrule
\end{tabular}}
\caption{Data format across arithmetic tasks for fine-tuning Qwen-Math model. Mathematical signs are removed for training minGPT. The meaning of the numbers in 2-digit CR: the number of heads is $A_1A_2$, the number of legs is $B_1B_2$, the number of chickens is $C_1C_2$, and the number of rabbits is ${C_3C_4}$. Abbreviations: GCD — greatest common divisor, CR — chicken-rabbit problem.}
\label{tab:arithmetic_dataformat} 
\end{table}

For the Qwen experiments, we fine-tune the open-source Qwen2.5-Math-1.5B\footnote{\url{https://huggingface.co/Qwen/Qwen2.5-Math-1.5B}} model using the official Hugging Face implementation. Training is performed with the AdamW optimizer, a learning rate of $1\times10^{-4}$, a batch size of 16, a gradient accumulation step of 4, and up to 30 epochs. One NVIDIA RTX 5090 GPU is used in this experiment. The software environment includes PyTorch 2.7.1, Hugging Face Transformers 4.52.4, CUDA 12.9, and Python 3.12.0 running on Ubuntu 24.04.2 LTS.

Table~\ref{tab:arithmetic_dataformat} outlines the input and output token formats used in each arithmetic task. Inputs are structured as digit sequences $A_1A_2\cdots B_1B_2\cdots$, as source tokens. The target sequence consists of the output digits $C_1C_2\cdots$, which the model is trained to predict. In the 4-digit Logarithm task, only one input number ($A_1A_2A_3A_4$) is provided. In contrast, the 2-digit Chicken-Rabbit task involves two output numbers: the number of chickens ($C_1C_2$) and the number of rabbits ($C_3C_4$).

All arithmetic datasets are programmatically generated using Python. For the GPT-based and Qwen-Math multi-task experiments, we enumerate all possible examples for each task and split them into 90\% training and 10\% testing data. For example, 2-digit and 3-digit multiplication contain 10,000 and 1,000,000 examples, respectively. For long-sequence Qwen-Math multiplication, the full data space grows by a factor of 100 whenever each operand gains one digit. Due to computational constraints, we generate the full data space for each digit length, uniformly sample 100,000 examples, and split them into 90\% training and 10\% testing data. Evaluation is based on fixed-iteration accuracy, where the iteration count is determined by the token order that achieves an accuracy plateau. Results are averaged over 3 runs with fixed random seeds for reproducibility. Randomness is controlled across model initialization, tokenizers, and data shuffling using \texttt{torch.manual\_seed} and \texttt{random.seed}.

\subsection{Multi-label Classification}\label{sec:B.2}
We fine-tune several small instruction-tuned LLMs from the GPT-2 \citep{radford2019language}, Llama-3.1, Llama-3.2 \citep{grattafiori2024llama3herdmodels}, and Qwen-2.5 \citep{qwen2025qwen25technicalreport} model families to assess scalability to larger models. GPT-2\footnote{\url{https://huggingface.co/openai-community/gpt2}} is fine-tuned for up to $40$ epochs using the AdamW optimizer, with a batch size of $8$ and a learning rate of $4 \times 10^{-5}$. Llama-3.1-8B-Instruct\footnote{\url{https://huggingface.co/meta-llama/Llama-3.1-8B-Instruct}}, Llama-3.2-1B-Instruct\footnote{\url{https://huggingface.co/meta-llama/Llama-3.2-1B-Instruct}}, Qwen2.5-3B-Instruct\footnote{\url{https://huggingface.co/Qwen/Qwen2.5-3B-Instruct}}, and Qwen2.5-1.5B-Instruct\footnote{\url{https://huggingface.co/Qwen/Qwen2.5-1.5B-Instruct}} models are fine-tuned for up to $20$ epochs using the AdamW optimizer and QLoRA \citep{NEURIPS2023_1feb8787} with 4-bit quantization, a rank of $16$, an alpha of $32$, a learning rate of $4 \times 10^{-7}$, and a batch size of $12$. These small LLMs are fine-tuned on the English and Chinese version of the ToxicComment dataset. The Chinese datasets are translated using Google Translate. Fine-tuning is conducted using two NVIDIA RTX A6000 GPUs. The software environment includes PyTorch 2.4.1, Hugging Face Transformers 4.45.0, CUDA 12.9, and Python 3.10.14 running on Ubuntu 20.04.6 LTS.

To standardize target sequence length, we retain four labels per dataset: marked as $C_1$, $C_2$, $C_3$, and $C_4$. In ToxicComment, \emph{toxic}, \emph{obscene}, \emph{insult}, and \emph{identity\_hate} are retained. In PaperAbstract, \emph{computer\_science}, \emph{physics}, \emph{mathematics}, and \emph{statistics} are retained. In GoEmotions, \emph{admiration}, \emph{amusement}, \emph{love}, and \emph{optimism} are retained. We remove samples with no positive label among the retained classes. After filtering, the train/test sizes of the three datasets are 13,687/1,044, 14,480/825, and 7,936/856.

\subsection{Text Generation}\label{sec:B.3}

GPT-2 and Llama-3.2-1B are continually pretrained on WikiText-2 using the official Hugging Face implementation. For GPT-2, we use the AdamW optimizer with a learning rate of $3\times10^{-5}$, 10 epochs, and a batch size of 4. For Llama-3.2-1B, we use the AdamW optimizer with a learning rate of $5\times10^{-5}$, 10 epochs, block size 1024, and batch size 4. A fixed random seed of 123 ensures reproducibility. Experiments are conducted on a single NVIDIA RTX 3090 GPU. The software environment includes PyTorch 2.6.0, Hugging Face Transformers 4.51.3, CUDA 12.9, and Python 3.9.21 on Ubuntu 20.04.6 LTS.

\begin{figure*}[t]
\centering
\includegraphics[width=1.0\textwidth]{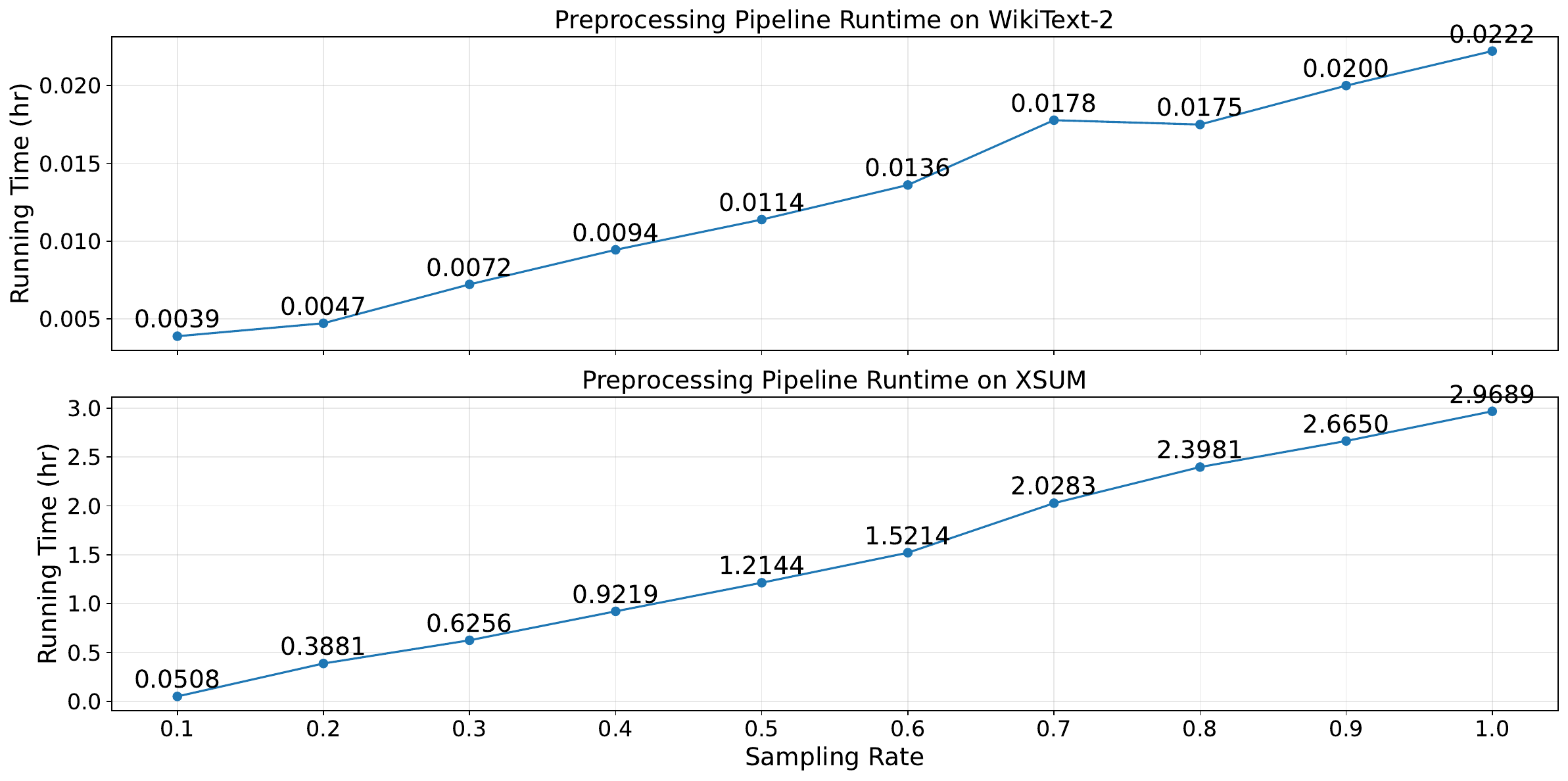}
\caption{Running time of the preprocessing pipeline on WikiText-2 and XSUM datasets across data sampling rates.}
\label{fig:preprocessing_time}
\end{figure*}

For summarization, Llama-3.2-1B-Instruct is fine-tuned on the full XSUM dataset. Training is performed for 1 epoch with the AdamW optimizer and QLoRA with 4-bit quantization, a rank of $16$, an alpha of $64$, a learning rate of $4 \times 10^{-5}$, a batch size of $1$, a gradient accumulation step of $4$, a weight decay of $1 \times 10^{-4}$, and a warmup ratio of $0.05$. This experiment is run on one NVIDIA RTX 5090 GPU. The environment includes PyTorch 2.7.1, Transformers 4.52.4, CUDA 12.9, and Python 3.12.0 on Ubuntu 24.04.2 LTS.

In our implementation, the dataset preprocessing process involves four stages: tokenization and lemmatization, joint-probability estimator training, MI scoring, and prefix-augmentation formatting. The total time complexity is $O(T+V^2)$, where T is the number of tokens in the dataset, and V is the number of lemmatized unique words. In practice, with an Intel(R) Xeon(R) w5-2455X CPU and a NVIDIA RTX 5090 GPU, it takes only 80 seconds to finish the preprocessing pipeline on WikiText-2, where V is 52257. 

To further assess the running time of the preprocessing pipeline under diverse data sizes, for both WikiText-2 and XSUM datasets, we measure the running time across data sampling rates, as shown in Figure \ref{fig:preprocessing_time}. Figure \ref{fig:preprocessing_time} shows that preprocessing is faster on WikiText-2 than on XSUM because WikiText-2 contains fewer samples (44,836 vs. 226,711) and shorter documents. On both datasets, runtime scales approximately linearly with dataset size, measured by the number of tokens.

\section{Experiments on GLUE Benchmark}\label{sec:C}

We use the \emph{GLUE benchmark}\footnote{\url{https://gluebenchmark.com/}} to evaluate the performance of the model fine-tuned on datasets preprocessed using different preprocessing strategies. GLUE, which stands for General Language Understanding Evaluation, is a popular benchmark designed to evaluate the performance of natural language understanding (NLU) systems. The compared strategies include NTP, TF-IDF, P(t|S), and Max(MI(S;t)).

\subsection{Experimental Setup}

We fine-tune GPT-2-small (137M parameters) on the following tasks in the GLUE benchmark: The Corpus of Linguistic Acceptability (CoLA), The Stanford Sentiment Treebank (SST-2), Microsoft Research Paraphrase Corpus (MRPC), Semantic Textual Similarity Benchmark (STS-B), Quora Question Pairs (QQP), Multi-Genre Natural Language Inference (MNLI), Stanford Question Answering (QNLI), Recognizing Textual Entailment (RTE), and Winograd Schema Challenge (WNLI). Only the diagnostic dataset is omitted.

We adapt the official Hugging Face Transformers GLUE example\footnote{\url{https://github.com/huggingface/transformers/blob/main/examples/pytorch/text-classification/run_glue.py}} with minor modifications. We use GPT-2-small, train for 6 epochs, set the learning rate to $2\times10^{-5}$, and use a batch size of 32. One NVIDIA RTX 5090 GPU is used in this experiment. The software environment includes PyTorch 2.7.1, Hugging Face Transformers 4.52.4, CUDA 12.9, and Python 3.12.0 running on Ubuntu 24.04.2 LTS.

\begin{table*}[htp!]\centering
\centering
\resizebox{\linewidth}{!}{
\begin{tabular}{llccccc}
    \toprule
    \textbf{Task} & \textbf{Metric(s)} & \textbf{NTP} & \textbf{TF-IDF} & \textbf{P($t|S$)} & \textbf{Max(MI($S$;$t$))}
    \\\toprule
    CoLA    & Matthew's Corr        & 0.4039 & 0.4411 & 0.4325 & \textbf{0.4803} \\\midrule
    SST-2   & Accuracy              & \textbf{0.9266} & 0.9197 & 0.9243 & \textbf{0.9266} \\\midrule
    MRPC    & F1 / Accuracy         & \textbf{0.8711}/0.8064 & 0.8703/\textbf{0.8137} & 0.8707/\textbf{0.8137} & 0.8527/0.7892 \\\midrule
    STS-B   & Pearson-Spearman Corr & \textbf{0.8518}/\textbf{0.8498} & 0.8444/0.8438 & 0.8465/0.8465 & 0.8391/0.8398 \\\midrule
    QQP     & F1 / Accuracy         & 0.8679/0.8994 & 0.8687/\textbf{0.9010} & 0.8663/0.8996 & \textbf{0.8688}/0.9008 \\\midrule
    MNLI    & Accuracy              & 0.8262 & 0.8222 & 0.8235 & \textbf{0.8288} \\\midrule
    QNLI    & Accuracy              & 0.8807 & \textbf{0.8843} & 0.8788 & 0.8794 \\\midrule
    RTE     & Accuracy              & 0.6606 & 0.6498 & 0.6426 & \textbf{0.6679} \\\midrule
    WNLI    & Accuracy              & 0.4930 & 0.5211 & 0.5352 & \textbf{0.5634} \\\midrule
    Average &                       & 0.7781 & 0.7817 & 0.7817 & \textbf{0.7864} \\
    \bottomrule
\end{tabular}
}
\caption{
    Comparison of preprocessing strategies on the GLUE benchmark using GPT-2. NTP denotes the standard unaugmented baseline; TF-IDF, P($t|S$), and Max(MI($S$;$t$)) use the prefix-augmentation format with different word-selection criteria.
}
\label{tab:glue_gerneration_res}
\end{table*}

\subsection{Results}

Table \ref{tab:glue_gerneration_res} demonstrates the model performance fine-tuned on the benchmark datasets across the compared strategies. Overall, Max(MI(S;$t$)) shows the highest average score across nine datasets (0.7864), with relative gains over NTP of 19\% on CoLA and 14\% on WNLI. However, the Max(MI(S;$t$)) strategy does not yield improvements in certain datasets, such as MRPC and STS-B, which are classified as similarity and paraphrase tasks in the GLUE benchmark.

\section{Dataset License and Anonymization}\label{sec:D}

ToxicComment\footnote{\url{https://huggingface.co/datasets/google/jigsaw_toxicity_pred}}: This dataset is released under the CC0 1.0 Universal (CC0 1.0) Public Domain Dedication. The data comprises Wikipedia comments labeled for toxic behavior, with personal identifiers removed to protect user privacy. PaperAbstract\footnote{\url{https://www.kaggle.com/datasets/blessondensil294/topic-modeling-for-research-articles}}: Available on Kaggle, this dataset contains abstracts of research articles intended for topic modeling. The dataset is provided for educational and research purposes, with sensitive information excluded to maintain anonymity. GoEmotions\footnote{\url{https://huggingface.co/datasets/google-research-datasets/go_emotions}} \citep{demszky-etal-2020-goemotions}: Distributed under the Apache License 2.0, this dataset includes carefully curated Reddit comments annotated for 27 emotion categories. The data has been anonymized to remove any references that could identify individual users, ensuring compliance with privacy standards. The WikiText-2\footnote{\url{https://huggingface.co/datasets/Salesforce/wikitext}} \citep{DBLP:journals/corr/MerityXBS16} dataset is distributed under the Creative Commons Attribution-ShareAlike 4.0 International (CC BY-SA 4.0) License. This license permits sharing and adaptation of the material, provided appropriate credit is given and any derivatives are distributed under the same license. Regarding anonymization, WikiText-2 is derived from verified "Good" and "Featured" Wikipedia articles. These articles are subject to Wikipedia's content policies, which generally exclude personal and sensitive information. XSUM\footnote{\url{https://huggingface.co/datasets/EdinburghNLP/xsum}} \citep{narayan-etal-2018-dont}: Released under the MIT License, the dataset contains 226,711 Wayback archived BBC articles ranging over almost a decade (2010 to 2017) and covering a wide variety of domains (e.g., News, Politics, Sports, Weather, Business, Technology, Science, Health, Family, Education, Entertainment, and Arts). GLUE \citep{wang-etal-2018-glue}: The primary GLUE tasks are built on and derived from existing datasets. We refer users to the original licenses accompanying each dataset.

\begin{figure*}[t]
\includegraphics[width=0.48\linewidth]{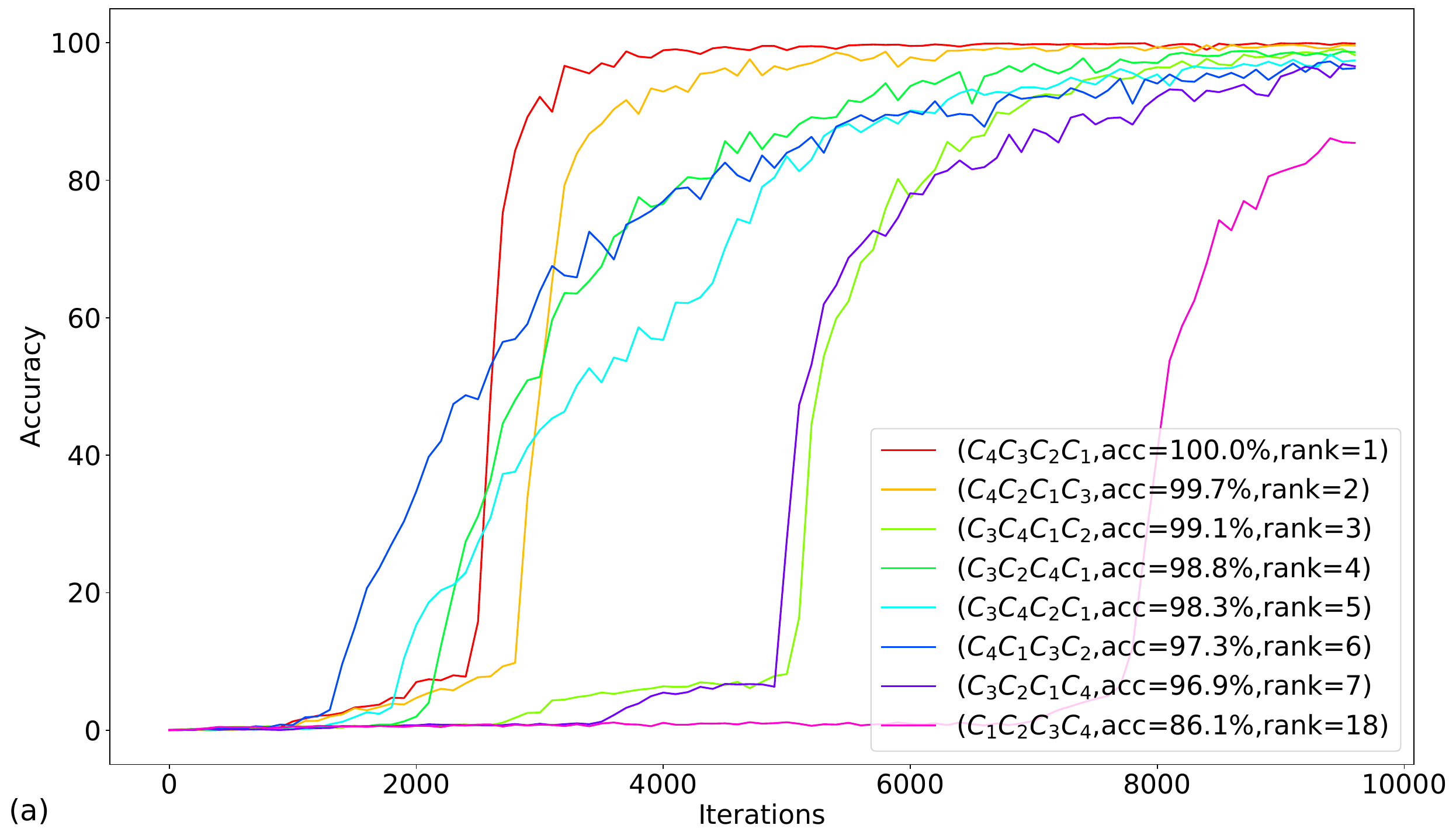} \hfill
\includegraphics[width=0.48\linewidth]{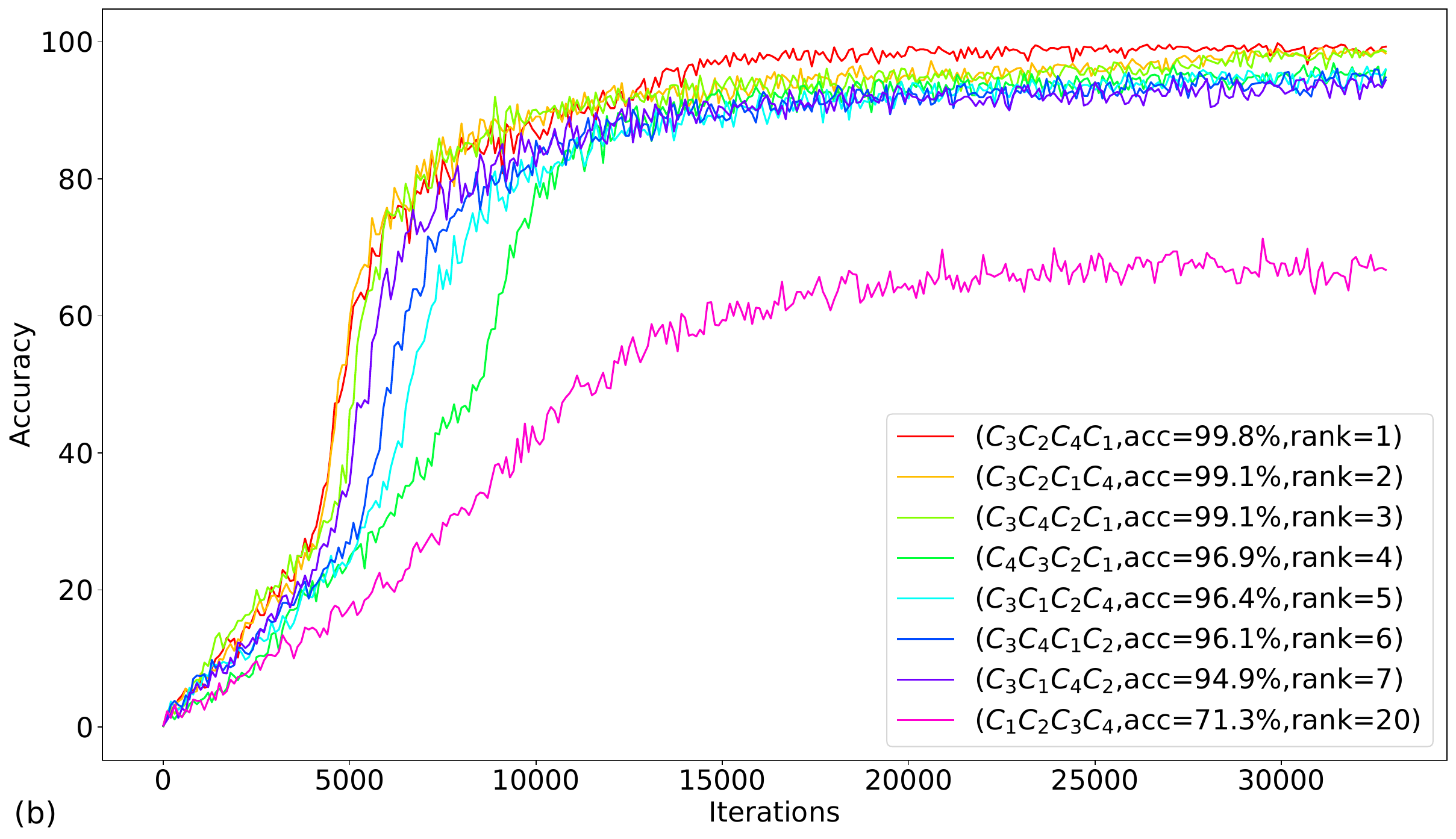}
\includegraphics[width=0.48\linewidth]{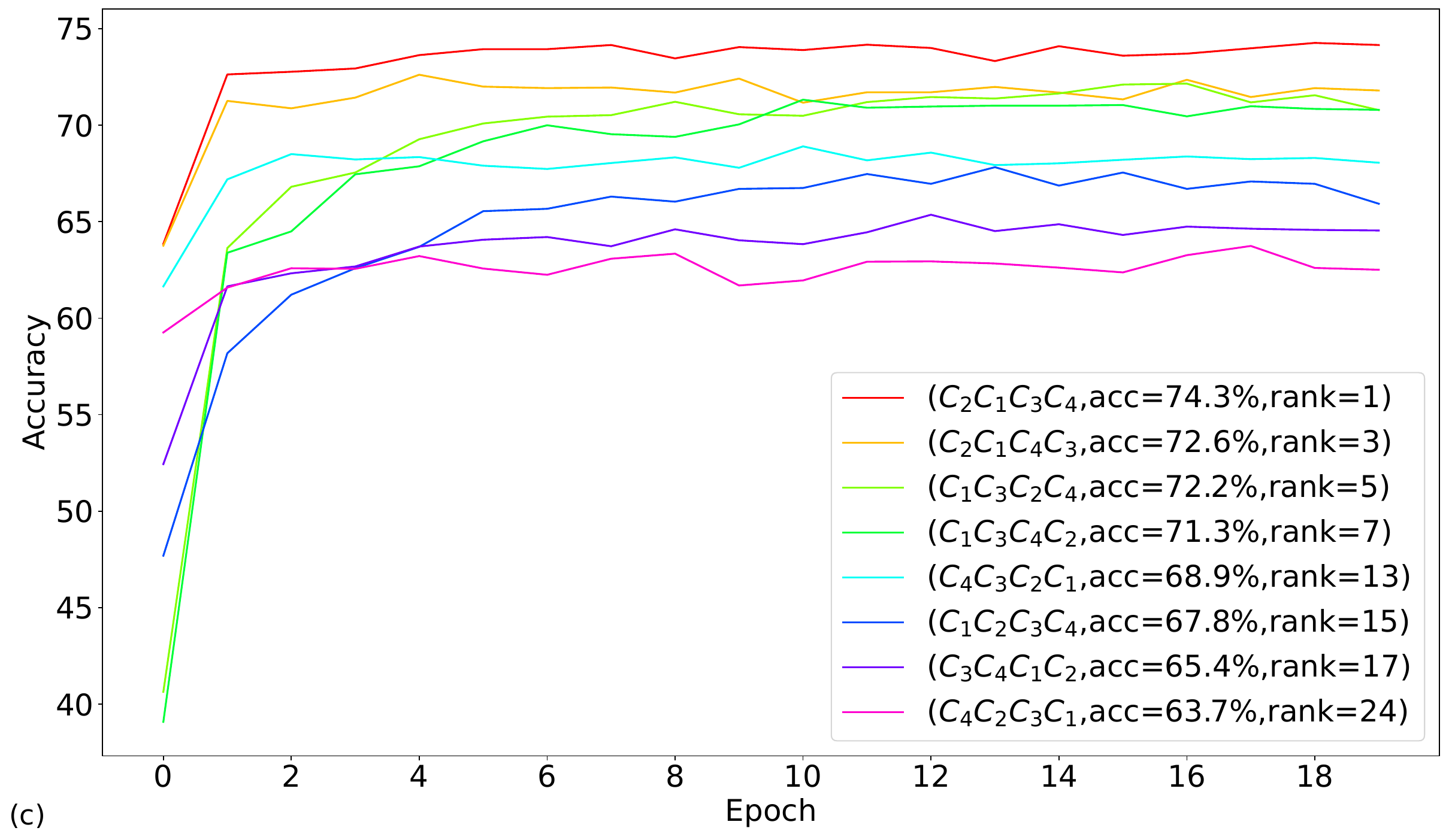} \hfill
\includegraphics[width=0.48\linewidth]{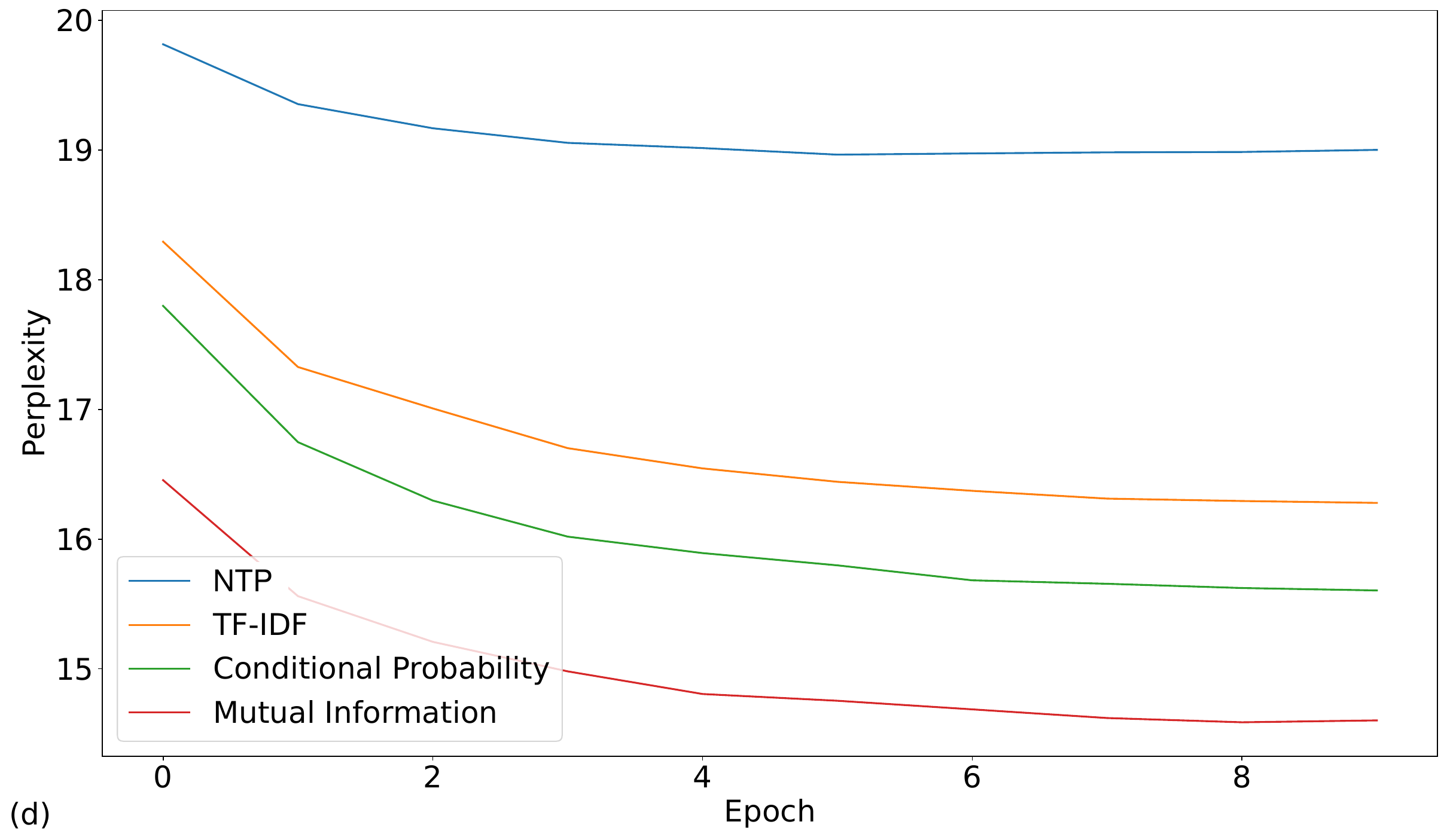}
  \caption {Training curves of different target token orders with significant differences (not displaying all permutations of target token orders due to complexity). (a) 3-digit Addition ${A_1A_2A_3 + B_1B_2B_3 = C_1C_2C_3C_4}$ on NanoGPT (0.09M parameters). (b) 2-digit Multiplication ${A_1A_2 \times B_1B_2 = C_1C_2C_3C_4}$ on GPT-2-mini (2.67M parameters). (c) Multi-label classification on Qwen-2.5-1.5B-Instruct using ToxicComment dataset, whose labels [$toxic, obscene, insult, identity\_hate$] are retained and marked as [${C_1,C_2,C_3,C_4}$]. Each curve represents a distinct token order. (d) Text generation on GPT-2-small (137M parameters) using WikiText-2 dataset. Abbreviations: Acc -- Maximal accuracy at fixed iteration, Rank -- the rank of Acc among all permutations of target token orders.}
  \label{fig:curves}
\end{figure*}

\vspace{0.24cm}
\section{Further Discussion: Connection to the Finding of Hallucination}\label{sec:E}
\vspace{0.24cm}

This study builds on previous research by demonstrating that inserting information-rich tokens at the beginning of sentences during pretraining enhances model training efficiency and reduces perplexity in text generation. Prior work suggests that MI-based token selection mitigates hallucination by reducing over-reliance on domain-specific token distributions \citep{van2022mutual}. Our findings align with this, showing that the Max(MI($S$;$t$)) strategy stabilizes token predictions and lowers perplexity. However, a key difference is that while the study applies pointwise MI adjustments during decoding, our method pre-orders tokens based on MI before training, suggesting that pretrain reordering can improve model efficiency without requiring real-time adjustments during generation. Previous research has explored TF-IDF for content weighting in summarization and text generation \citep{van2022mutual}. Our findings reinforce this by showing that TF-IDF-based copying improves perplexity but is slightly less effective than Max(MI($S$;$t$)), likely because MI accounts for both input context and uncertainty. Additionally, exposure bias studies suggest that randomized or default ordering strategies degrade performance due to cascading errors \citep{ranzato2015sequence,wang2020exposure}. Our findings confirm that unstructured copying (e.g., Random word) does not contribute to meaningful improvements, supporting the necessity of principled reordering strategies. Unlike methods that adjust token probabilities dynamically during inference, our approach leverages reordering at the training stage, making it computationally efficient. Future work could explore hybrid methods that integrate pretraining-based MI reordering with uncertainty-aware or TF-IDF-based decoding strategies, leveraging both pre-ordering and dynamic token selection. Additionally, investigating task-specific adaptations of MI reordering across diverse NLP domains could provide deeper insights into optimizing sequence modeling strategies.

\end{document}